\documentclass[journal]{IEEEtran}

\usepackage{bm}
\usepackage{amsmath}
\usepackage{graphicx}
\usepackage{amssymb}
\usepackage{multirow}
\usepackage{array}
\usepackage{color}
\usepackage{epsfig}
\usepackage{algorithmic}
\usepackage{graphics} 
\usepackage{epsfig} 

\allowdisplaybreaks

\renewcommand{\algorithmicrequire}
{{\em Input:}}  
\renewcommand{\algorithmicensure}
{{\em Output:}}  

\newtheorem{theorem}{Theorem}

\newcommand{\lfteqn}{\begin{eqnarray} \begin{array}{lllllll}}
\newcommand{\ndeqn}{\end{array} \nonumber \end{eqnarray}}
\newcommand{\Lfteqn}{\begin{eqnarray} \begin{array}{lllllll}}
\newcommand{\Ndeqn}{\end{array}  \end{eqnarray}}

\newcommand{\cN}{\cal{N}}
\newcommand{\cS}{\cal{S}}

\title{Convergence of a New Learning Algorithm}

\author{Feng Lin, {\em Fellow, IEEE} 
\thanks{Feng Lin is with the Department of Electrical and Computer
Engineering, Wayne State University, Detroit, MI 48202, U.S.A.
E-mail: flin@wayne.edu. }
        }

\begin{document}

\maketitle \thispagestyle{empty} \pagestyle{empty}

\begin{abstract}

A new learning algorithm proposed by Brandt and Lin for neural
network \cite{brandt1996supervised, lin2020supervised} has been
shown to be mathematically equivalent to the conventional
back-propagation learning algorithm, but has several advantages
over the back-propagation algorithm, including
feedback-network-free implementation and biological plausibility.
In this paper, we investigate the convergence of the new
algorithm. A necessary and sufficient condition for the algorithm
to converge is derived. A convergence measure is proposed to
measure the convergence rate of the new algorithm. Simulation
studies are conducted to investigate the convergence of the
algorithm with respect to the number of neurons, the connection
distance, the connection density, the ratio of
excitatory/inhibitory synapses, the membrane potentials, and the
synapse strengths.


\end{abstract}

\section{Introduction}

Since (artificial) neural networks were introduced more than 70
years ago, many results have been obtained
\cite{rochester1956tests, rosenblatt1958perceptron,
	hopfield1982neural, mcclelland1986parallel}, despite two
``winters'', during which the interests and funding for neural
networks were significantly reduced. Since 2005, there have been
renewed interests in neural networks in the form of deep learning
\cite{bengio2013representation, schmidhuber2015deep,
	lecun2015deep, krizhevsky2017imagenet}. Deep learning has been
successfully applied to many practical problems, including face
recognition \cite{ding2015robust, masi2018deep, guo2019survey},
speech recognition \cite{graves2013speech, deng2013new,
	hannun2014deep, nassif2019speech}, object detection
\cite{szegedy2013deep, erhan2014scalable, pathak2018application,
	zhao2019object}, and game playing \cite{mnih2013playing,
	gibney2016google, silver2016mastering, lanctot2017unified}.

The main characteristics of neural networks is their ability to
learn. Several learning algorithms have been proposed in the past
\cite{stephen1990perceptron, caruana2006empirical,
	ayodele2010types, japkowicz2011evaluating}. Among them, the
conventional back-propagation algorithm (abbreviated as the B-P
algorithm in the rest of the paper) is probably most popular and
has many advantages \cite{rumelhart1986learning,
	pineda1987generalization, hecht1992theory,
	chauvin2013backpropagation}. Learning using the B-P algorithm
requires a feedback neural network to back-propagate errors. This
feedback neural network must have the same topology and connection
weights as the feed-forward neural network being learned. This
requirement makes the realization of the B-P algorithm in
biological neural networks unlikely. Hence, the use of the B-P
algorithm is inconsistent with the claim that (artificial) neural
networks mimic the learning in biological neural networks.

To overcome this inconsistence and to remove the requirement of
feedback neural networks, a new learning algorithm was proposed by
Brandt and Lin (abbreviated as the B-L algorithm in the rest of
the paper) in \cite{brandt1996supervised, lin2020supervised}. The
B-L algorithm is mathematically equivalent to the B-P algorithm.
However, unlike the back-propagation algorithm, it calculates the
derivatives of strengths/weights (for learning) of dendritic
synapses/connections based on the strengths and their
derivatives of axonic synapses/connections of the same neuron.
Hence, information needed for learning is available locally in
each neuron and there is no need to use a feedback neural network
to back-propagate errors.

The B-L algorithm has several advantages over the B-P algorithm.
One notable advantage is that it makes the learning mathematically
equivalent to the B-P algorithm biologically plausible. Hence,
artificial neural networks may indeed mimic the learning in
biological neural networks.

In this paper, we investigate the convergence of the B-L
algorithm. We use a biological neural network model in our
investigation. The connection among neurons in the network can be
arbitrary and without restriction. In particular, the model covers
both hierarchical neural networks and non-hierarchical neural
networks. If a neural network is not hierarchical, then the B-L
algorithm is given by a set of implicit functions. We derive a
necessary and sufficient condition for the set of implicit
functions to have a unique solution, which is expressed as a
matrix being invertible. If the condition is satisfied, we
investigate the convergence of the B-L algorithm to this solution.
We show that the convergence is determined by the eigenvalues of a
matrix that depends on the number of neurons, the connection
distance, the connection density, the ratio of
excitatory/inhibitory synapses, the membrane potentials, and the
synapse strengths.

We use the maximal absolute value of eigenvalues of the matrix as
the convergence measure so that smaller convergence measure means
faster convergence. We use simulations to derive relations between
the convergence measure and the following parameters: (1) the
number of neurons, (2) the connection distance, (3) the connection
density, (4) the ratio of excitatory/inhibitory synapses, (5) the
membrane potentials, and (6) the synapse strengths, respectively.
We randomly generate neural networks with different parameter
values and then determine their convergence measures. The
following results are obtained. (1) Convergence measure increases
as the number of neurons increases. (2) Convergence measure
increases as the connection distance increases. (3) Convergence
measure increases as the connection density increases. (4)
Convergence measure is smallest when the ratio of
excitatory/inhibitory synapses is 1. (5) Convergence measure
deceases as the membrane potential increases. (6) Convergence
measure increases as the synapse strength increases.

The paper is organized as follows. We first present a biological
neural network model and review the B-L learning algorithm. A
necessary and sufficient condition for the B-L algorithm to have a
unique solution is then derived. A necessary and sufficient
condition for the B-L algorithm to converge is also derived. A
convergence measure is introduced to study the convergence rate.
Finally, we investigate convergence of the B-L algorithm with
respect to various parameters described above.

\section{Biological Neural Network Model}

Biological neural networks are usually not strictly layered. To
model neural networks with general connections, the following
model for neural networks is proposed in
\cite{brandt1996supervised, lin2020supervised}. This general model
includes both hierarchical neural networks and non-hierarchical
neural networks as special cases. The model also includes
biological parameters so that their impacts on the convergence of
the B-L algorithm can be evaluated.

Denote the number of neurons in a neural network by $N$. Enumerate
the neurons in the neural network as ${\cN} = \{ 1, 2, ..., N \}$.
Hence, a neuron in the neural network can be denoted as $n \in
\cN$. Similarly, enumerate the synapses in the neural network as
${\cS} = \{ 1, 2, ..., S \}$ and denote a synapse in the neural
network as $s \in \cS$.

For each neuron $n \in \cN$, denote the set of its dendritic
synapses as $D_n$ and the set of axonic synapses as $A_n$. For
each synapse $s \in \cS$, denote its presynaptic neuron as $pre_s$
and its postsynaptic neuron as $post_s$. We use $\epsilon_s$ to
indicate whether the synapse $s$ is excitatory (+1) or inhibitory
(-1). Therefore, for each synapse $s \in \cS$, the triplet
($pre_s$, $post_s$, $\epsilon_s$) specifies the presynaptic
neuron, the postsynaptic neuron, and whether the synapse is
excitatory or inhibitory, respectively. It is also clear that for
all $s \in D_n$, $n=post_s$ and for all $s \in A_n$, $n=pre_s$.
These relations are shown in Figure 1.

\begin{figure}[htb] \label{Figure1}
	\centering
	\includegraphics[keepaspectratio=true,angle=0,width=0.45\textwidth]{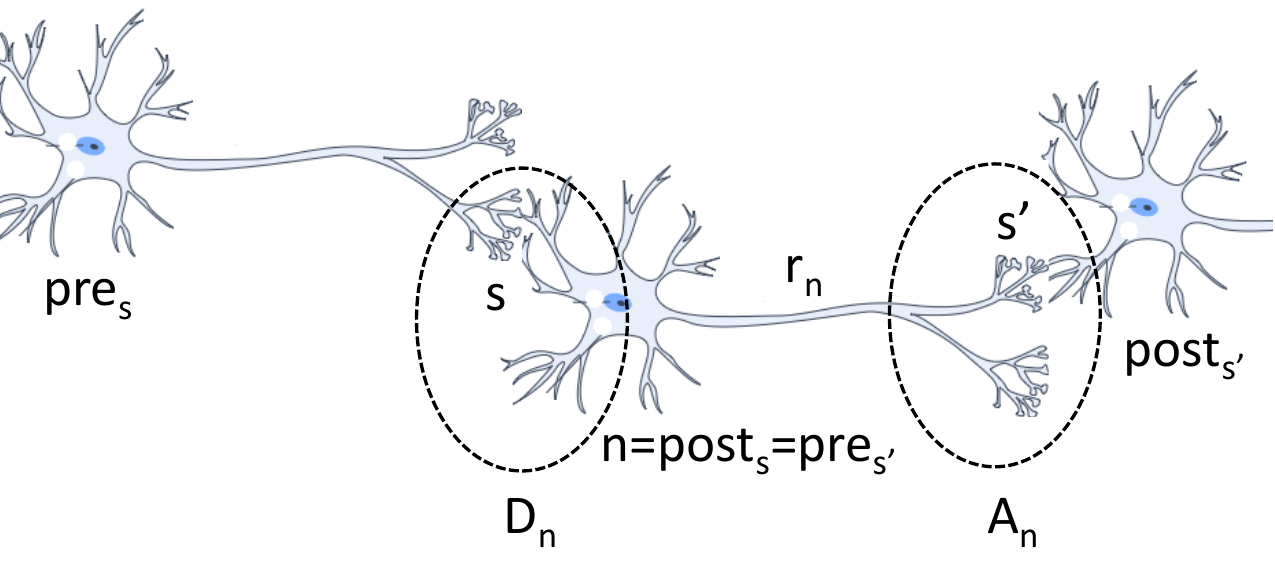}
	\caption{Labels for neurons and synapses in a neural network.}
\end{figure}

For each synapse $s \in \cS$, $w_s$ denotes its strength. For each
neuron $n \in \cN$, $p_n$ denotes its membrane potential and $r_n$
denotes its firing rate, which is the output of the neuron.

In a neuron $n \in \cN$, signals are transmitted from $D_n$ to
$A_n$ in the following order. (1) Postsynaptic potentials are
generated following activation of a synapse by neurotransmitters.
(2) Postsynaptic potentials are spatially and temporally
integrated at the soma. (3) Action potentials is triggered along
the axon hillock and propagated along the axon. (4)
Neurotransmitters is released from presynaptic terminals.

In a neuron $n \in \cN$, the firing rate is defined as the
reciprocal of the interspike interval. The synapse strength $w_s$
of a synapse is proportional to the quantity of neurotransmitters
released when a spike arrives at the synapse. We assume that the
short-time average of the postsynaptic potential is proportional
to the product of the synapse strength $w_s$ and the presynaptic
firing rate $r_{pre_s}$. The membrane potential $p_n$ is the sum
of the postsynaptic potentials as follows.

\begin{equation}\label{Eq1}
	p_n = \sum _{s \in D_n} \epsilon _s \eta_s w_s r_{pre_s} - \theta
	_n,
\end{equation}
where $\eta_s$ is the decremental conduction coefficient of $s$
\cite{de1959decremental}; and $\theta _n$ is the offset of the
membrane potential of $n$ \cite{montaigne1984offset}.

The firing rate $r_n$ is a sigmoidal function of its membrane
potential as follows.

\begin{equation}\label{Eq2}
	r_n = \rho _n \sigma (\lambda _n p_n),
\end{equation}
where $\rho_n$ is the maximal firing rate of $n$; and $\lambda _n
$ is the measure of the steepness of the sigmoidal characteristic
of $n$.

\section{The B-L Learning Algorithm}

Different learning algorithms have been proposed in the literature
to learn the strengths of synapses so that some error is
minimized. The following least square error is widely used for
learning.
\begin{equation} \label{least}
	\begin{split}
		E=\frac{1}{2} \sum _{n \in {\cal{O}}} (r_n - \bar{r}_n)^2,
	\end{split}
\end{equation}
where ${\cal{O}}$ is the set of output neurons and $\bar{r}_n$ is
the desired/target output of an output neuron.

Among different learning algorithms, the B-P algorithm is very
popular. On the other hand, the B-L algorithm is mathematically
equivalent to the B-P algorithm and is give by
\begin{equation} \label{Eq3}
	\begin{split}
		\dot{w_s} = & \alpha_s \epsilon _s r_{pre_s} \rho _{post_s}
		\lambda_{post_s} \sigma'(\lambda_{post_s} p_{post_s}) \\
		& (\frac{1}{r_{post_s}} \sum _{s' \in A_{post_s}}
		\frac{\eta_{s'}}{\alpha_{s'}} w_{s'} \dot{w}_{s'} - \gamma
		(r_{post_s} - \bar{r}_{post_s})) .
	\end{split}
\end{equation}

This algorithm reduces the error $E$ monotonically because the
following equation is satisfied \cite{brandt1996supervised,
	lin2020supervised}.
\begin{equation}\label{Eq5}
	\dot{w_s} = - \alpha _s \frac{\gamma}{\eta_s}\frac{dE}{dw_s}.
\end{equation}

It is shown in \cite{brandt1996supervised, lin2020supervised} that
the B-L algorithm has the following properties.

\begin{enumerate}
	\item
	The B-L algorithm and the B-P algorithm are mathematically
	equivalent. Hence, they have the same performance and can be used
	in same applications.
	\item
	The B-L algorithm is easy to implement because it does not require
	a feedback network for error back-propagation. A
	feedback-network-free implementation is given in
	\cite{lin2020supervised}.
	\item
	Since it is unlikely that the requirement of feedback network can
	be met in biological neural networks, removal of feedback network
	makes the learning according to the B-L algorithm is much more
	plausible in biological neural networks.
	\item
	In the B-L algorithm, the information needed for dendritic
	synapses to learn is implicit in the weights and their derivatives
	of axonic synapses of the same neuron.
	\item
	The B-L algorithm allows a phaseless learning by processing
	information asynchronously and concurrently without the needs of a
	feed-forward phase and a feedback phase.
	\item
	In a layered neural network, all layers will have the same or
	similar structures using the B-L algorithm. Hence, the B-L
	algorithm can be implemented in Simulink and other implementation
	platforms using same or similar blocks.
	\item
	The B-L algorithm provides a much simpler implementation on
	silicon without complex wiring among neurons, as there is no need
	to build a separate feedback network and connect it to the
	feed-forward network.
	\item
	An adaptive neuron can be designed using the B-L algorithm as an
	identical and standard unit. These adaptive neurons can then be
	interconnected arbitrarily. This provides the potential for
	designing neural networks with dynamically reconfigurable
	topologies.
	\item
	The B-L algorithm are more fault-tolerant since all feedback and
	connections are local. Hence, failures of some neurons will not
	significantly affect the entire neural network.
\end{enumerate}

\section{Existence of the Solution to the B-L Learning Algorithm}

A neural network is called hierarchical if there exists a partial
order $\leq$ on ${\cal{N}}$ such that, for $n_1 , n_2 \in
{\cal{N}}$, if $n_1 \leq n_2$, then there exist no axonic synapses
of $n_2$ with $n_1$, that is, $D_{n_1} \cap A_{n_2} = \emptyset$.
This partial order ensures that the output of any neuron $n$ is an
explicit function of the outputs of some preceding neurons $n'$
satisfying $n' \leq n \wedge n' \not= n$. Hence, the output of
each neuron is always well-defined and unique. Layered neural
networks are special cases of hierarchical neural networks.

It is shown in \cite{brandt1996supervised, lin2020supervised} that
the B-L algorithm can be used for both hierarchical and
non-hierarchical networks. If the network is non-hierarchical,
Equation (\ref{Eq3}) gives only implicit functions of $\dot{w_s}$.

To ensure the existence of these implicit functions, let us write
Equation (\ref{Eq3}) in the following matrix form.
\begin{align*}
	& \left[ \begin{array}{l}
		\dot{w_1} \\
		\dot{w_2}\\
		... \\
		\dot{w_S}
	\end{array} \right] =
	\left[ \begin{array}{llll}
		g_{11} & g_{12} & ... & g_{1S} \\
		g_{21} & g_{22} & ... & g_{2S} \\
		... \\
		g_{S1} & g_{S2} & ... & g_{SS}
	\end{array} \right]
	\left[ \begin{array}{l}
		\dot{w_1} \\
		\dot{w_2}\\
		... \\
		\dot{w_S}
	\end{array} \right] +
	\left[ \begin{array}{l}
		h_1 \\
		h_2 \\
		... \\
		h_S
	\end{array} \right]
\end{align*}

Denote the above matrix equation as
\begin{equation}\label{Eq6}
	\dot{w} = G \dot{w} + h,
\end{equation}
where $\dot{w}$ and $h$ are column vectors of $S$ dimension, and
$G=[g_{ij}]$ is an $S \times S$ matrix with
$$
g_{ij} = \alpha_i \epsilon _i r_{pre_i} \rho _{post_i}
\lambda_{post_i} \sigma'(\lambda_{post_i} p_{post_i})
\frac{\eta_{j} w_{j} }{r_{post_i}\alpha_{j}},
$$
if $j \in A_{post_i}$ and $g_{ij} = 0$, otherwise.

For the implicit functions of $\dot{w}$ to exist, it is necessary
and sufficient that the determinant
$$
|I-G| \not= 0,
$$
where $I$ is the identity matrix of the dimension $S \times S$.

Under this assumption, the solution of $\dot{w}$, denoted by
$\dot{w}^*$ is given by
$$
\dot{w}^* = (I-G)^{-1}h.
$$

Note that for hierarchical neural networks, the condition of
$|I-G| \not= 0$ is always satisfied. To prove this, let us
enumerate the synapses according to the enumeration of neurons
such that
$$
j \leq i \Rightarrow post_j \leq post_i.
$$
One such enumeration is to enumerate all synapses in $D_1$, then
all synapses in $D_2$, ..., etc. Under such an enumeration, we
have
\begin{align*}
	& j \leq i \\
	\Rightarrow & post_j \leq post_i \\
	\Rightarrow & D_{post_j} \cap A_{post_i} = \emptyset \\
	& (\mbox{by the definition of hierarchical networks}) \\
	\Rightarrow & j \not\in  A_{post_i} \\
	& (\mbox{because } j \in D_{post_j}) \\
	\Rightarrow & g_{ij} =0 \\
	& (\mbox{by the definition of } g_{ij}) .
\end{align*}
Hence, $G$ is a triangular matrix with all elements on the
diagonal equal 0. Therefore, $|I-G| = 1 \not= 0$.

\section{Convergence of the B-L Learning Algorithm}

In the rest of the paper, we assume that, for a given neural
network, the solution $\dot{w}^*$ exists, that is, $|I-G| \not=
0$. We further assume that the neural network is deep or rich
enough so that learning can be accomplished, that is, using a
gradient descent algorithm, $\lim _{t \rightarrow \infty} r_n (t)
= \bar{r}_n$, for all $n \in {\cal{O}}$. These two assumptions are reasonable, because, without them, learning cannot be achieved by any learning algorithms.

We investigate the convergence of the B-L algorithm, that is,
whether $(\exists \bar{w}) \lim _{t \rightarrow \infty} w (t) =
\bar{w}$, where $\bar{w}$ is a constant column vector representing
the final value of $w$. Clearly, $(\exists \bar{w}) \lim _{t
	\rightarrow \infty} w (t) = \bar{w}$ if and only if $\lim _{t
	\rightarrow \infty} \dot{w} (t) = 0$. Therefore, the B-L algorithm
converges if and only if $\lim _{t \rightarrow \infty} \dot{w} (t)
= 0$.

To check whether $\lim _{t \rightarrow \infty} \dot{w} (t) = 0$,
let $\dot{w} (t)$ be calculated recursively using Equation
(\ref{Eq6}) as follows. Denote the sample of $\dot{w} (t)$
($h(t)$, respectively) at time $t_k= k \Delta$ by $\dot{w}(k)$
($h(k)$, respectively), where $k=0,1,2,...$ and $\Delta$ is the
sampling period. Let
$$
\dot{w}(k+1) = G \dot{w}(k) + h(k) .
$$
Then,
\begin{align*}
	& \lim _{t \rightarrow \infty} \dot{w} (t) = 0 \\
	\Leftrightarrow & \lim _{k \rightarrow \infty} \dot{w} (k) = 0 \\
	\Leftrightarrow & (\forall \alpha \in eig(G)) \ |\alpha| <1 \wedge
	\lim _{k \rightarrow \infty} h(k) = 0,
\end{align*}
where $eig(G)$ denotes the set of eigenvalues of $G$.

Since $\lim _{t \rightarrow \infty} r_n (t) = \bar{r}_n$, for all
$n \in {\cal{O}}$ by assumption, $\lim _{k \rightarrow \infty} h
(k) = 0$ is true. Therefore, $\lim _{t \rightarrow \infty} \dot{w}
(t) = 0$ if and only if $ (\forall \alpha \in eig(G)) \ |\alpha|
<1. $

We summarize the above results in the following theorem.

\begin{theorem}
	The B-L algorithm converges if and only if
	\begin{equation}\label{Eq8}
		(\forall \alpha \in eig(G)) \ |\alpha| <1. \\
	\end{equation}
	
\end{theorem}

We use the maximum of the absolute values of all eigenvalues of
$G$ as the measurement of convergence, that is, the convergence
measure is given by
$$
CM=\max_{\alpha \in eig(G)} \ |\alpha| .
$$
Hence, the smaller $CM$ is, the faster the B-L Algorithm
converges.

\section{Simulation Study}

Using Theorem 1 and the convergence measure $CM$ defined above, we can investigate convergence of the B-L algorithm under different parameters by ``simulating'' the convergence
measure $CM$ for those parameters.  In the simulations below, we consider the following parameters.

$N$ = the number of neurons.

$B$ = the connection distance, that is, neuron $n$ may be
connected to neurons $n' \in \{n-B, n-B+1, ..., n, n+1, ..., n+B
\}$.

$C$ = the connection density, that is, the percentage of possible
connections that are actually connected.

$D$ = the percentage of synapses that are excitatory ($\epsilon =
1$). Hence, $D/(1-D)$ is the ratio of excitatory/inhibitory
synapses.

$\hat{p}_n$ = the normalized membrane potential, which includes
$\lambda$.

$\hat{w}_s$ = the normalized synapse strengths, which includes
$\rho$ and $\eta$.

For each set of parameters, we run $R=20$ trials, each trial
randomly generates a neural network with random connections and
with the parameters $\hat{p}_n$ and $\hat{w}_s$ perturbed by
0\%-10\% to ensure the robustness of the results. We then take the
maximum of $CM$ over $R$ trials as the $CM$ for this set of
parameters.

Since the objective of the simulations is to investigate the
impacts of different parameters on the convergence of the B-L
algorithm, the relative values (rather than the absolute values)
of the parameters are of importance. Hence, we start with the
following set of parameters
\lfteqn
N=30, & B=20, & C=0.5, \\
D=0.5, & \hat{p}_n=5, & \hat{w}_s = 25.
\ndeqn
that gives a reasonable $CM \approx 0.8$. In the simulations, we
are not concerned with the absolute values $p_n$ and $w_s$ or
their units. We investigate the impact of each parameter on the
convergence of the B-L algorithm as described below. The impact of several parameters on the convergence of the B-L algorithm is the sum of the impacts of the parameters involved.

\subsection{Convergence vs the number of neurons}

To investigate the convergence of the B-L algorithm vs the number
of neurons, we simulate neural networks with the number of neurons
ranges from 5 to 50 (note that the nominal value is 30). All other
parameters are at their nominal values given above. The results
are shown in Figure 2.

\begin{figure}[htb] \label{Figure2}
	\centering
	\includegraphics[keepaspectratio=true,angle=0,width=0.5\textwidth]{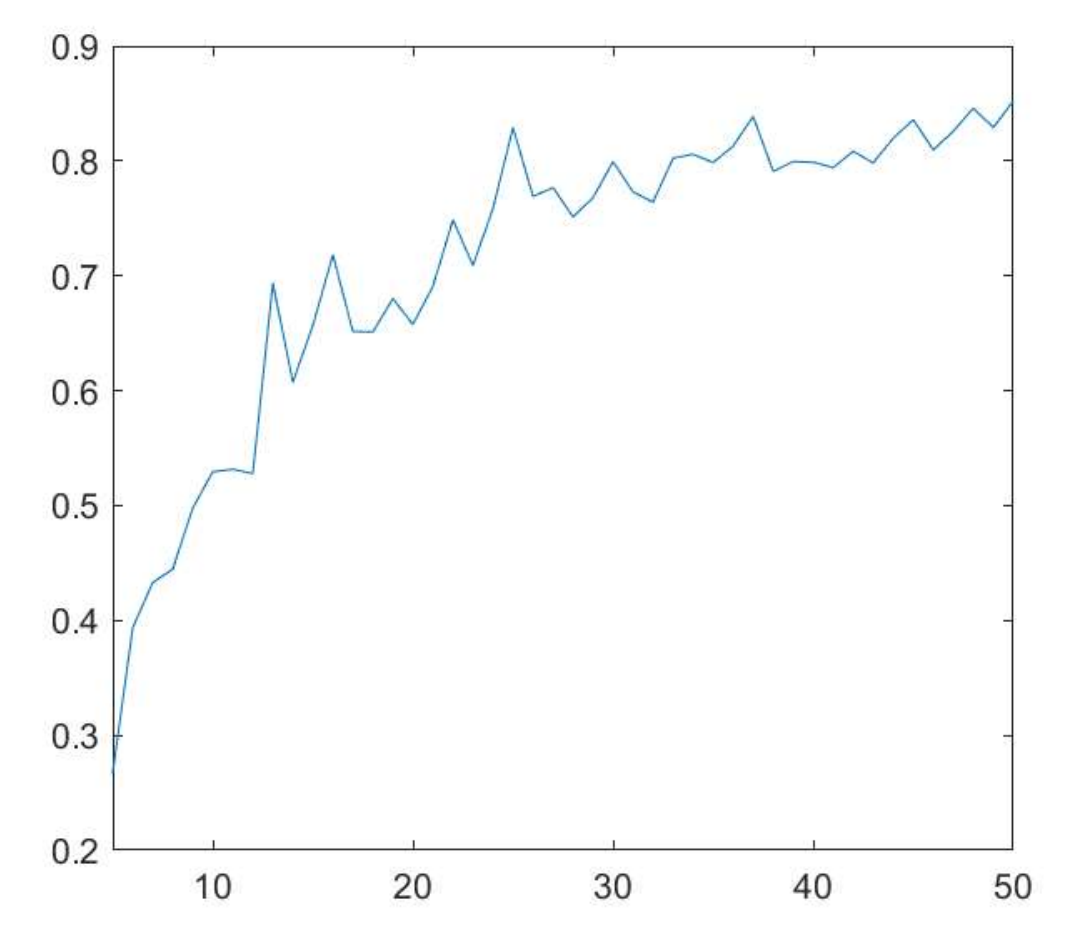}
	\caption{Convergence measure $CM$ (vertical axis) of the B-L algorithm vs number of neurons $N$ (horizontal axis)}
\end{figure}

The simulations show that the convergence measure increases as the
number of neurons increases. In other words, the B-L algorithm
converges faster in small neural networks than in large neural
networks. On the other hand, the rate of increase in the the
convergence measure reduces as the number of neurons increases.
The B-L algorithm still converges when $N=50$.

\subsection{Convergence vs the connection distance}

To investigate the convergence of the B-L algorithm vs the
connection distance, we simulate neural networks with the
connection distance ranges from 1 to 30 (the maximum). All other
parameters are at their nominal values given above. The results
are shown in Figure 3.

\begin{figure}[htb] \label{Figure3}
	\centering
	\includegraphics[keepaspectratio=true,angle=0,width=0.5\textwidth]{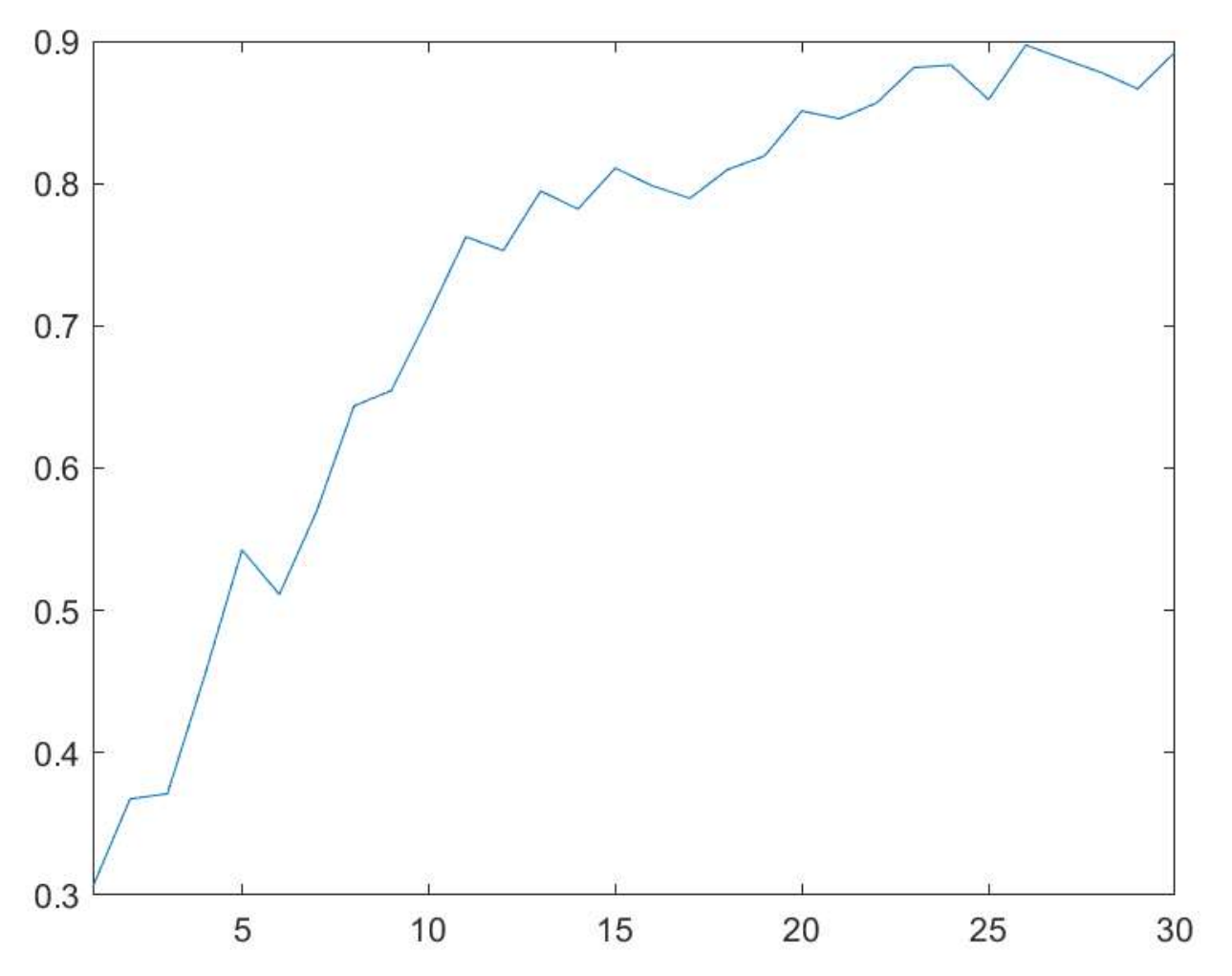}
	\caption{Convergence measure $CM$ (vertical axis) of the B-L algorithm vs connection distance $B$ (horizontal axis)}
\end{figure}

The simulations show that the convergence measure increases as the
connection distance increases and the increase is less than the
linear increase. Hence, the B-L algorithm converges faster in
neural networks with local connections than in neural networks
with long-distance connections.

\subsection{Convergence vs the connection density}

To investigate the convergence of the B-L algorithm vs the
connection density, we simulate neural networks with the
connection density ranges from 0.1 to 0.9. All other parameters
are at their nominal values given above. The results are shown in
Figure 4.

\begin{figure}[htb] \label{Figure4}
	\centering
	\includegraphics[keepaspectratio=true,angle=0,width=0.5\textwidth]{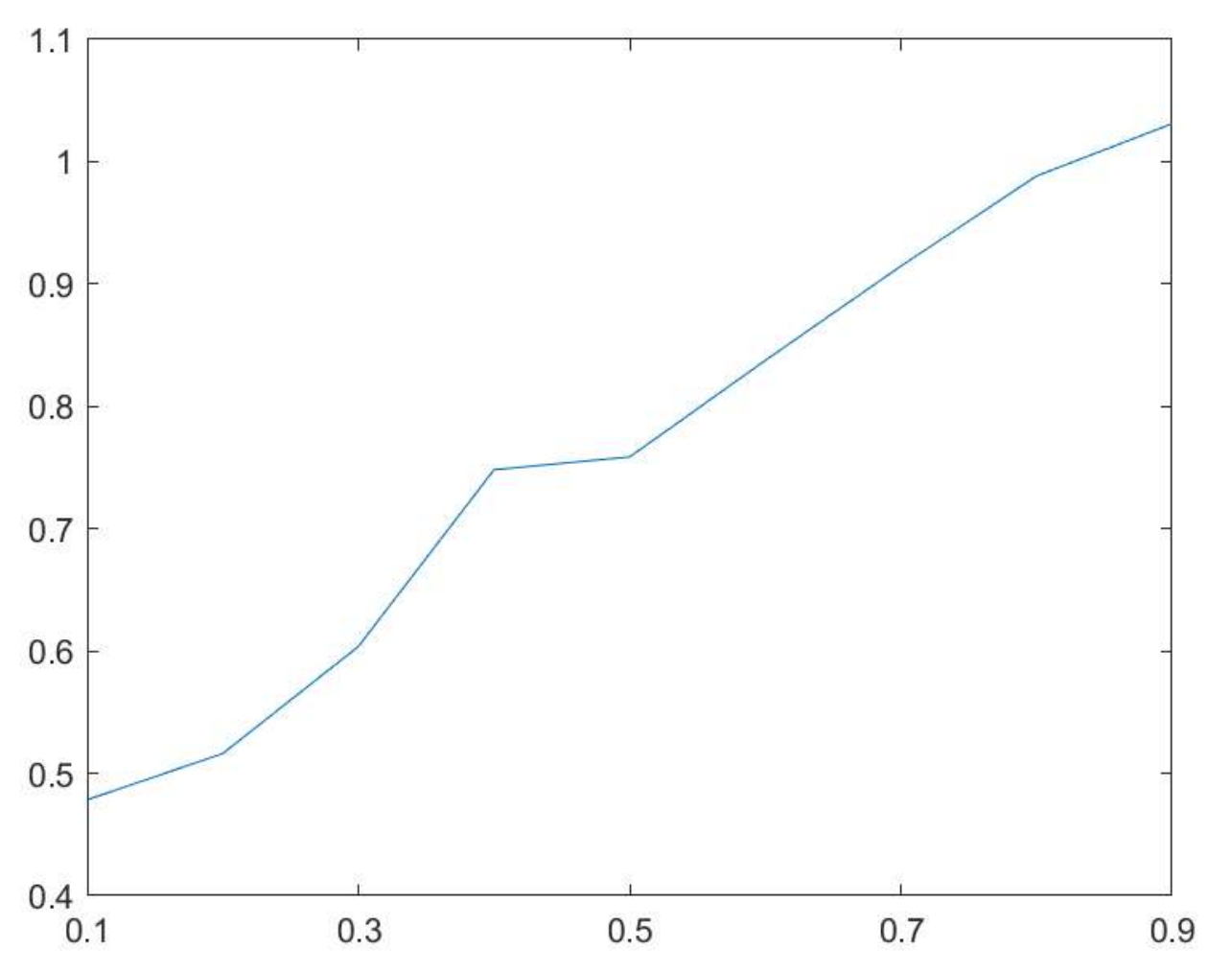}
	\caption{Convergence measure $CM$ (vertical axis) of the B-L algorithm vs connection density $C$ (horizontal axis)}
	\caption{Convergence of the B-L algorithm vs connection density.}
\end{figure}

The simulations show that the convergence measure increases almost
linearly as the connection density increases. Therefore, the B-L
algorithm converges slower in neural networks with more dense
connections.

\subsection{Convergence vs the ratio of excitatory/inhibitory synapses}

To investigate the convergence of the B-L algorithm vs the ratio
of excitatory/inhibitory synapses, we simulate neural networks
with the percentage of synapses that are excitatory ranges from
0.1 to 0.9. All other parameters are at their nominal values given
above. The results are shown in Figure 5.

\begin{figure}[htb] \label{Figure5}
	\centering
	\includegraphics[keepaspectratio=true,angle=0,width=0.5\textwidth]{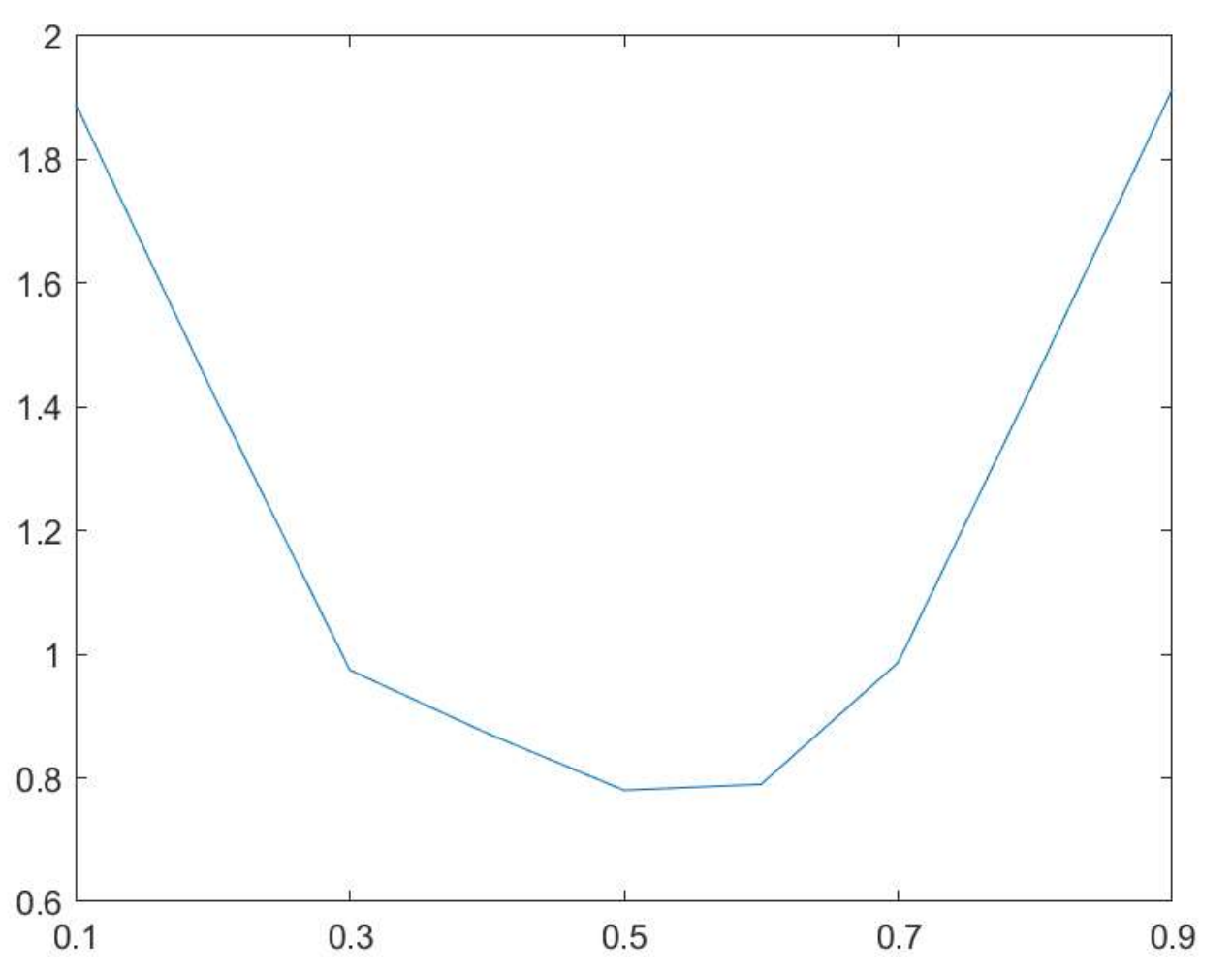}
	\caption{Convergence measure $CM$ (vertical axis) of the B-L algorithm vs percentage of synapses that are excitatory $D$ (horizontal axis)}
\end{figure}

The simulations show that the convergence measure is smallest when
the percentage of synapses that are excitatory is 0.5, that is,
when the ratio of excitatory/inhibitory synapses is 1.

Biological neural networks have both excitatory and inhibitory
synapses, which is good for the convergence of the B-L algorithm.
Having only excitatory synapses or only inhibitory synapses may
cause the B-L algorithm not converge.

\subsection{Convergence vs the membrane potentials}

To investigate the convergence of the B-L algorithm vs the
membrane potentials, we simulate neural networks with membrane
potential ranges from 4 to 10. All other parameters are at their
nominal values given above. The results are shown in Figure 6.

\begin{figure}[htb] \label{Figure6}
	\centering
	\includegraphics[keepaspectratio=true,angle=0,width=0.5\textwidth]{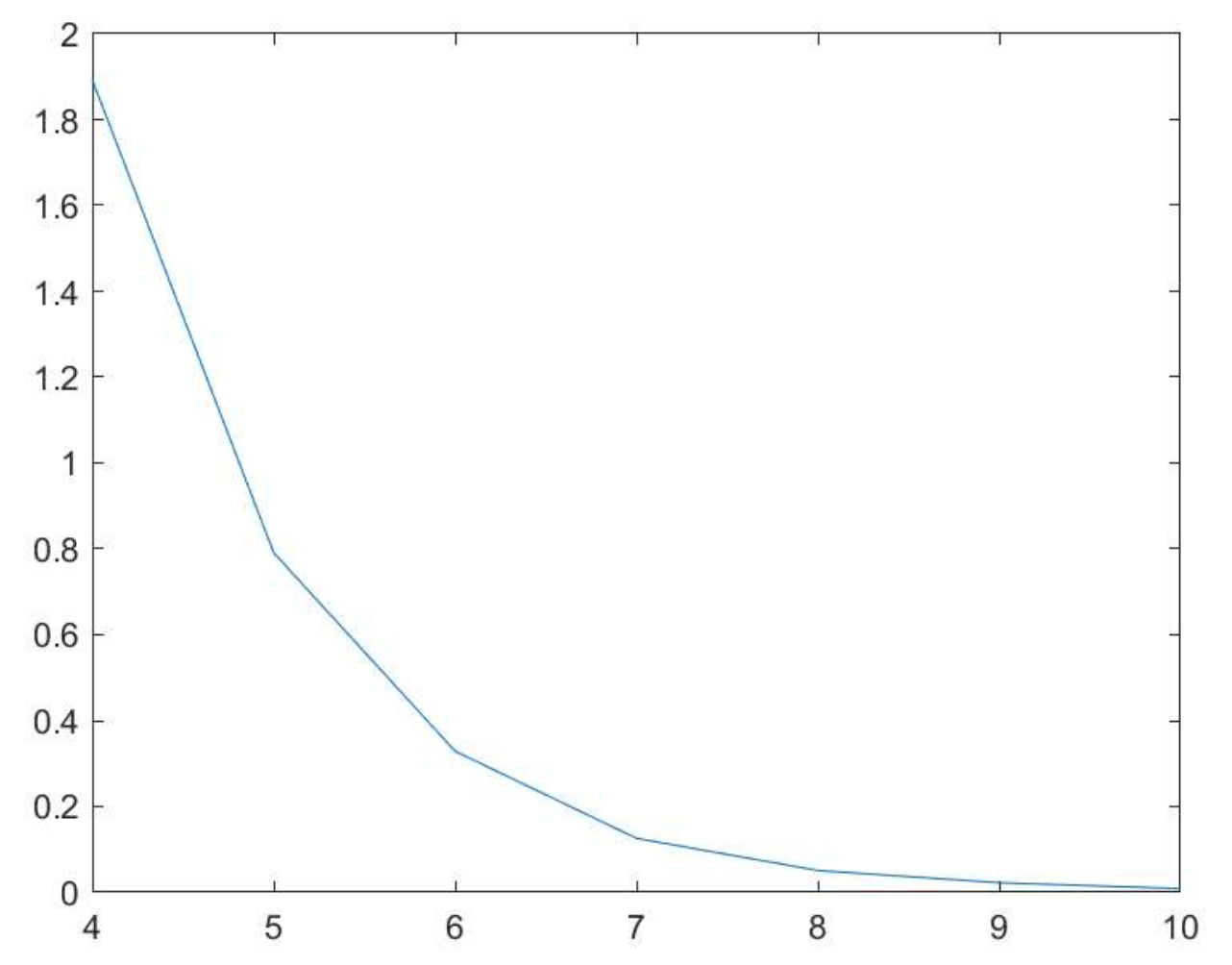}
	\caption{Convergence measure $CM$ (vertical axis) of the B-L algorithm vs  membrane potential $\hat{p}_n$ (horizontal axis)}
\end{figure}

The simulations show that the convergence measure decreases almost
exponentially as the membrane potentials increases. Therefore, the
B-L algorithm converges faster in neural networks with high
membrane potentials. Hence, it is important to maintain membrane
potentials at certain level in biological neural networks.

\subsection{Convergence vs the synapse strengths}

To investigate the convergence of the B-L algorithm vs the synapse
strengths, we simulate neural networks with the synapse strengths
ranges from 20 to 30. All other parameters are at their nominal
values given above. The results are shown in Figure 7.

\begin{figure}[htb] \label{Figure7}
	\centering
	\includegraphics[keepaspectratio=true,angle=0,width=0.5\textwidth]{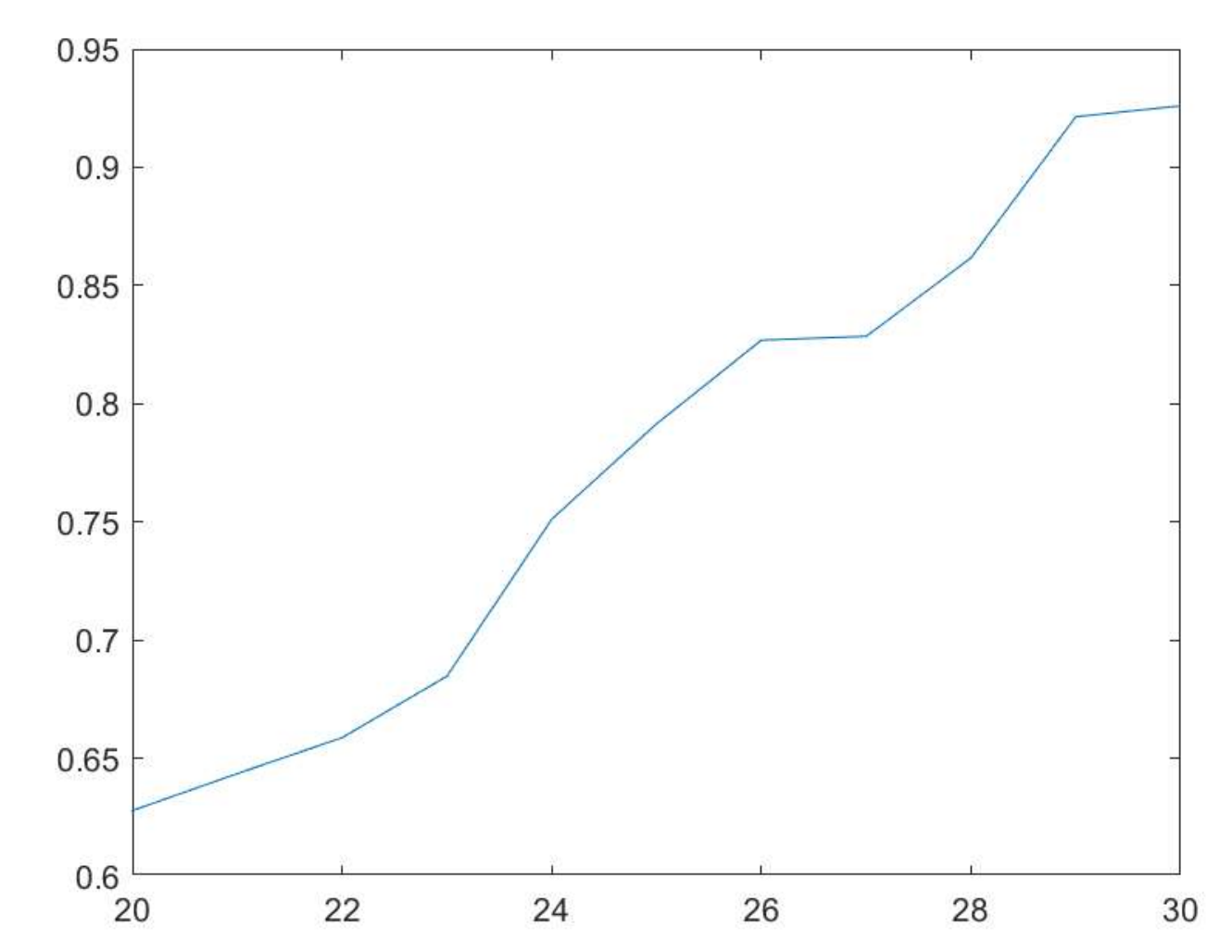}
	\caption{Convergence measure $CM$ (vertical axis) of the B-L algorithm vs synapse strengths $\hat{w}_s$ (horizontal axis)}
\end{figure}

The simulations show that the convergence measure increases as the
synapse strengths increases more or less linearly. In other words,
the B-L algorithm converges slower in neural networks with large
synapse strengths. It is possible that some mechanisms exist in
biological neural networks to ensure synapse strengths are not too
large.

\section{Conclusions}

We investigate the convergence of the B-L algorithm in this paper.
We first derive a necessary and sufficient condition for the B-L
algorithm to converge. We then propose a convergence measure to
study the convergence rate of the B-L algorithm. Using
simulations, we show how the convergence measure is related to
various parameters of neural networks. The parameters include (1)
the number of neurons, (2) the connection distance, (3) the
connection density, (4) the ratio of excitatory/inhibitory
synapses, (5) the membrane potentials, and (6) the synapse
strengths. The simulations shows that the membrane potentials have
the biggest impact on the convergence of the B-L algorithm.

%

\begin{thebibliography}{10}

\bibitem{brandt1996supervised}
R.~D. Brandt and F.~Lin, ``Supervised learning in neural networks without
feedback network,'' in {\em Intelligent Control, 1996., Proceedings of the
1996 IEEE International Symposium on}, pp.~86--90, IEEE, 1996.

\bibitem{lin2020supervised}
F.~Lin, ``Supervised learning in neural networks: Feedback-network-free
implementation and biological plausibility,'' {\em IEEE Transactions on
Neural Networks and Learning Systems}, 2021.

\bibitem{rochester1956tests}
N.~Rochester, J.~Holland, L.~Haibt, and W.~Duda, ``Tests on a cell assembly
theory of the action of the brain, using a large digital computer,'' {\em IRE
Transactions on information Theory}, vol.~2, no.~3, pp.~80--93, 1956.

\bibitem{rosenblatt1958perceptron}
F.~Rosenblatt, ``The perceptron: a probabilistic model for information storage
and organization in the brain.,'' {\em Psychological review}, vol.~65, no.~6,
p.~386, 1958.

\bibitem{hopfield1982neural}
J.~J. Hopfield, ``Neural networks and physical systems with emergent collective
computational abilities,'' {\em Proceedings of the national academy of
sciences}, vol.~79, no.~8, pp.~2554--2558, 1982.

\bibitem{mcclelland1986parallel}
J.~L. McClelland, D.~E. Rumelhart, P.~R. Group, {\em et~al.}, ``Parallel
distributed processing,'' {\em Explorations in the Microstructure of
Cognition}, vol.~2, pp.~216--271, 1986.

\bibitem{bengio2013representation}
Y.~Bengio, A.~Courville, and P.~Vincent, ``Representation learning: A review
and new perspectives,'' {\em IEEE transactions on pattern analysis and
machine intelligence}, vol.~35, no.~8, pp.~1798--1828, 2013.

\bibitem{schmidhuber2015deep}
J.~Schmidhuber, ``Deep learning in neural networks: An overview,'' {\em Neural
networks}, vol.~61, pp.~85--117, 2015.

\bibitem{lecun2015deep}
Y.~LeCun, Y.~Bengio, and G.~Hinton, ``Deep learning,'' {\em nature}, vol.~521,
no.~7553, pp.~436--444, 2015.

\bibitem{krizhevsky2017imagenet}
A.~Krizhevsky, I.~Sutskever, and G.~E. Hinton, ``Imagenet classification with
deep convolutional neural networks,'' {\em Communications of the ACM},
vol.~60, no.~6, pp.~84--90, 2017.

\bibitem{ding2015robust}
C.~Ding and D.~Tao, ``Robust face recognition via multimodal deep face
representation,'' {\em IEEE Transactions on Multimedia}, vol.~17, no.~11,
pp.~2049--2058, 2015.

\bibitem{masi2018deep}
I.~Masi, Y.~Wu, T.~Hassner, and P.~Natarajan, ``Deep face recognition: A
survey,'' in {\em 2018 31st SIBGRAPI conference on graphics, patterns and
images (SIBGRAPI)}, pp.~471--478, IEEE, 2018.

\bibitem{guo2019survey}
G.~Guo and N.~Zhang, ``A survey on deep learning based face recognition,'' {\em
Computer Vision and Image Understanding}, vol.~189, p.~102805, 2019.

\bibitem{graves2013speech}
A.~Graves, A.-r. Mohamed, and G.~Hinton, ``Speech recognition with deep
recurrent neural networks,'' in {\em 2013 IEEE international conference on
acoustics, speech and signal processing}, pp.~6645--6649, IEEE, 2013.

\bibitem{deng2013new}
L.~Deng, G.~Hinton, and B.~Kingsbury, ``New types of deep neural network
learning for speech recognition and related applications: An overview,'' in
{\em 2013 IEEE international conference on acoustics, speech and signal
processing}, pp.~8599--8603, IEEE, 2013.

\bibitem{hannun2014deep}
A.~Hannun, C.~Case, J.~Casper, B.~Catanzaro, G.~Diamos, E.~Elsen, R.~Prenger,
S.~Satheesh, S.~Sengupta, A.~Coates, {\em et~al.}, ``Deep speech: Scaling up
end-to-end speech recognition,'' {\em arXiv preprint arXiv:1412.5567}, 2014.

\bibitem{nassif2019speech}
A.~B. Nassif, I.~Shahin, I.~Attili, M.~Azzeh, and K.~Shaalan, ``Speech
recognition using deep neural networks: A systematic review,'' {\em IEEE
Access}, vol.~7, pp.~19143--19165, 2019.

\bibitem{szegedy2013deep}
C.~Szegedy, A.~Toshev, and D.~Erhan, ``Deep neural networks for object
detection,'' {\em Advances in neural information processing systems},
vol.~26, pp.~2553--2561, 2013.

\bibitem{erhan2014scalable}
D.~Erhan, C.~Szegedy, A.~Toshev, and D.~Anguelov, ``Scalable object detection
using deep neural networks,'' in {\em Proceedings of the IEEE conference on
computer vision and pattern recognition}, pp.~2147--2154, 2014.

\bibitem{pathak2018application}
A.~R. Pathak, M.~Pandey, and S.~Rautaray, ``Application of deep learning for
object detection,'' {\em Procedia computer science}, vol.~132,
pp.~1706--1717, 2018.

\bibitem{zhao2019object}
Z.-Q. Zhao, P.~Zheng, S.-t. Xu, and X.~Wu, ``Object detection with deep
learning: A review,'' {\em IEEE transactions on neural networks and learning
systems}, vol.~30, no.~11, pp.~3212--3232, 2019.

\bibitem{mnih2013playing}
V.~Mnih, K.~Kavukcuoglu, D.~Silver, A.~Graves, I.~Antonoglou, D.~Wierstra, and
M.~Riedmiller, ``Playing atari with deep reinforcement learning,'' {\em arXiv
preprint arXiv:1312.5602}, 2013.

\bibitem{gibney2016google}
E.~Gibney, ``Google ai algorithm masters ancient game of go,'' {\em Nature
News}, vol.~529, no.~7587, p.~445, 2016.

\bibitem{silver2016mastering}
D.~Silver, A.~Huang, C.~J. Maddison, A.~Guez, L.~Sifre, G.~Van Den~Driessche,
J.~Schrittwieser, I.~Antonoglou, V.~Panneershelvam, M.~Lanctot, {\em et~al.},
``Mastering the game of go with deep neural networks and tree search,'' {\em
nature}, vol.~529, no.~7587, pp.~484--489, 2016.

\bibitem{lanctot2017unified}
M.~Lanctot, V.~Zambaldi, A.~Gruslys, A.~Lazaridou, K.~Tuyls, J.~P{\'e}rolat,
D.~Silver, and T.~Graepel, ``A unified game-theoretic approach to multiagent
reinforcement learning,'' in {\em Advances in neural information processing
systems}, pp.~4190--4203, 2017.

\bibitem{stephen1990perceptron}
I.~Stephen, ``Perceptron-based learning algorithms,'' {\em IEEE Transactions on
neural networks}, vol.~50, no.~2, p.~179, 1990.

\bibitem{caruana2006empirical}
R.~Caruana and A.~Niculescu-Mizil, ``An empirical comparison of supervised
learning algorithms,'' in {\em Proceedings of the 23rd international
conference on Machine learning}, pp.~161--168, 2006.

\bibitem{ayodele2010types}
T.~O. Ayodele, ``Types of machine learning algorithms,'' {\em New advances in
machine learning}, vol.~3, pp.~19--48, 2010.

\bibitem{japkowicz2011evaluating}
N.~Japkowicz and M.~Shah, {\em Evaluating learning algorithms: a classification
perspective}.
\newblock Cambridge University Press, 2011.

\bibitem{rumelhart1986learning}
D.~E. Rumelhart, G.~E. Hinton, and R.~J. Williams, ``Learning representations
by back-propagating errors,'' {\em nature}, vol.~323, no.~6088, pp.~533--536,
1986.

\bibitem{pineda1987generalization}
F.~J. Pineda, ``Generalization of back-propagation to recurrent neural
networks,'' {\em Physical review letters}, vol.~59, no.~19, p.~2229, 1987.

\bibitem{hecht1992theory}
R.~Hecht-Nielsen, ``Theory of the backpropagation neural network,'' in {\em
Neural networks for perception}, pp.~65--93, Elsevier, 1992.

\bibitem{chauvin2013backpropagation}
Y.~Chauvin and D.~E. Rumelhart, {\em Backpropagation: theory, architectures,
and applications}.
\newblock Psychology Press, 2013.

\bibitem{de1959decremental}
R.~L. De~N{\'o} and G.~Condouris, ``Decremental conduction in peripheral nerve.
integration of stimuli in the neuron,'' {\em Proceedings of the National
Academy of Sciences of the United States of America}, vol.~45, no.~4, p.~592,
1959.

\bibitem{montaigne1984offset}
K.~Montaigne and W.~F. Pickard, ``Offset of the vacuolar potential of characean
cells in response to electromagnetic radiation over the range 250 hz-250
khz,'' {\em Bioelectromagnetics: Journal of the Bioelectromagnetics Society,
The Society for Physical Regulation in Biology and Medicine, The European
Bioelectromagnetics Association}, vol.~5, no.~1, pp.~31--38, 1984.

\end{thebibliography}

\end{document}